\documentclass{article}
\usepackage[preprint]{colm2026_conference}

\usepackage{microtype}
\usepackage{graphicx}
\usepackage{booktabs}
\usepackage{amsmath}
\usepackage{amssymb}
\usepackage{mathtools}
\usepackage{tikz}
\usepackage{placeins}
\usepackage{xcolor}
\usepackage{hyperref}
\usepackage{url}
\usepackage{lineno}

\usetikzlibrary{arrows.meta,calc,positioning,backgrounds,decorations.markings}

\definecolor{darkblue}{rgb}{0, 0, 0.5}
\hypersetup{colorlinks=true, citecolor=darkblue, linkcolor=darkblue, urlcolor=darkblue}

\newcommand{\E}{\mathbb{E}}
\newcommand{\CE}{\operatorname{CE}}
\newcommand{\PPL}{\operatorname{PPL}}
\newcommand{\Kaiyue}[1]{}

\title{Training Hybrid Block Diffusion Language Models with Partial Bidirectionality}

\author{
Pranshu Chaturvedi \\
Stanford University \\
\texttt{pranshu@cs.stanford.edu}
\And
Parth Shroff \\
Stanford University \\
\texttt{pshroff@cs.stanford.edu}
\And
Tarun Suresh \\
Stanford University \\
\texttt{tsuresh@stanford.edu}
\And
Hangoo Kang \\
Stanford University \\
\texttt{hangook@stanford.edu}
\And
Kaiyue Wen \\
Stanford University \\
\texttt{kaiyuew@stanford.edu}
}

\begin{document}

\ifcolmsubmission
\linenumbers
\fi

\maketitle

\begin{abstract}
High-throughput long-context generation is one of the central challenges for
large language models. Generation is typically memory-bandwidth-bound rather
than compute-bound: each decoding step must stream the accumulated key/value
(KV) cache from memory, so bandwidth demand grows with context length while
only one token is emitted. Two parallel approaches have therefore emerged: reducing memory access
with efficient attention variants and linear-time mixers such as Mamba, or
increasing parallel computation by generating blocks of tokens at once. However,
technical challenges arise when combining these two ideas. Earlier hybrid diffusion
models such as DiffuMamba use bidirectional Mamba
mixing, including a reverse-direction scan relative to causal generation. This
reverse scan needs to scan the entire sequence, so its states are not
prefix-only and cannot be precisely reused as a cache even when diffusion is
performed block by block. We propose a BDLM Mamba--attention
hybrid that addresses this challenge by restricting the reverse Mamba scan to
the active denoising block, which enables exact caching across blocks. In an
87M-parameter DCLM sweep,
BDLM Mamba-H achieves the best C4-en validation perplexity
compared to BDLM attention and full-sequence baselines. At 350M parameters, it
remains competitive with BDLM attention. For long-context inference,
BDLM Mamba-H reaches 19.7x the throughput of full-sequence DiffuMamba-H at 65K
tokens and 3.7x the throughput of BDLM attention at 262K, showing that
Mamba hybrids are a potential long-context diffusion
architecture.
\end{abstract}

\section{Introduction}


Serving language models at long context remains largely a memory-bandwidth
problem. In autoregressive decoding, every generated token reads the accumulated
key/value (KV) cache, so bytes moved grow with context length while only one
token is emitted. Two lines of work address this pressure from different
directions. Hybrid autoregressive models reduce cache traffic by replacing many
attention layers with linear-time recurrent or state-space mixers such as
Mamba~\citep{gu2023mamba,dao2024mamba2,qin2024lightningattention,kimi2025kimilinear,li2025minimax01,ling2025ringlinear}.
Masked diffusion language models instead generate many tokens in parallel by
iterative denoising~\citep{austin2021d3pm,hoogeboom2021argmax,sahoo2024mdlm,nie2025llada}.
Block diffusion language models (BDLMs) make diffusion cacheable by generating
blocks left to right: completed prefix blocks become fixed conditioning, and
the current block is denoised in parallel~\citep{arriola2025blockdiffusion}.

DiffuMamba combines diffusion with Mamba and shows strong long-context
inference throughput for full-sequence denoisers~\citep{singh2026diffumamba}.
Its fully bidirectional Mamba scan, however, includes a reverse-direction pass
over the whole denoising window. Those reverse states depend on the active
tokens and are therefore not reusable prefix states. We study a native BDLM
Mamba--attention hybrid, BDLM Mamba-H, that restricts the reverse-direction
Mamba scan to the active denoising block. Completed prefix blocks are stored as
attention KV states for attention layers and as forward Mamba boundary states
for Mamba layers. The all-block objective constructs these same prefix caches
during training and lets downstream block losses train them directly, while the
active block keeps local bidirectional mixing.

In this work we explore whether cacheable BDLM Mamba hybrids can retain competitive
validation quality while improving long-context inference throughput. We also
describe timestep-conditioning factorization that applies to only the current active block for BDLM
Mamba-H, so timestep modulation can be applied without making clean-prefix
caches depend on the reverse diffusion step (Appendix~\ref{app:timestep}).
Our contributions are as follows.

\begin{figure}[!t]
  \centering
  \resizebox{0.58\columnwidth}{!}{\begin{tikzpicture}[
  x=1cm,
  y=1cm,
  font=\scriptsize,
  box/.style={
    draw=black,
    rounded corners=2pt,
    line width=0.55pt,
    align=center,
    minimum height=0.42cm,
    inner xsep=3.2pt,
    inner ysep=1.8pt
  },
  pblock/.style={box, fill=blue!10, minimum width=0.78cm},
  curblock/.style={box, fill=orange!13, minimum width=0.92cm},
  futblock/.style={box, fill=gray!10, text=gray!65!black, minimum width=0.78cm},
  op/.style={box, fill=green!12, minimum width=1.22cm},
  scanbox/.style={op, minimum width=1.62cm},
  cachebox/.style={op, minimum width=1.62cm},
  den/.style={box, fill=yellow!15, minimum width=1.52cm},
  time/.style={box, fill=gray!12, minimum width=0.68cm},
  loss/.style={box, fill=purple!8, minimum width=0.92cm},
  fwd/.style={-{Latex[length=1.35mm]}, line width=0.5pt},
  opt/.style={-{Latex[length=1.35mm]}, dashed, line width=0.5pt, draw=black},
  grad/.style={-{Latex[length=1.35mm]}, dashed, line width=0.55pt, draw=green!45!black},
  muted/.style={line width=0.45pt, draw=gray!55}
]
  \path[use as bounding box] (-0.55,-1.55) rectangle (6.08,3.22);

  \node[align=center] at (1.06,2.98) {Clean prefix $c^{(k)}$};
  \node[align=center] at (3.25,2.98) {Current\\block};
  \node[align=center, text=gray!65!black] at (4.91,2.98) {Future blocks\\masked out};

  \node[pblock] (b1) at (0.00,2.34) {$b_1^\star$};
  \node[pblock] (b2) at (1.06,2.34) {$b_2^\star$};
  \node[pblock] (b3) at (2.12,2.34) {$b_3^\star$};
  \node[curblock] (bk) at (3.25,2.34) {$\tilde b_t^{(k)}$};
  \node[futblock] (bf1) at (4.38,2.34) {$b_{t+1}^\star$};
  \node[futblock] (bf2) at (5.44,2.34) {$b_{t+2}^\star$};

  \node[scanbox] (scan) at (1.06,1.34) {Mamba Prefix\\scan};
  \node[cachebox] (cache) at (1.06,0.48) {$\operatorname{Cache}(c)$};
  \node[den] (denoise) at (3.25,0.48) {Active block\\denoiser};
  \node[time] (time) at (4.90,0.48) {$t$};
  \node[loss] (loss) at (3.25,-0.36) {CE loss};

  \draw[fwd] (b1.south) -- ($(scan.north west)!0.20!(scan.north east)$);
  \draw[fwd] (b2.south) -- ($(scan.north west)!0.50!(scan.north east)$);
  \draw[fwd] (b3.south) -- ($(scan.north west)!0.80!(scan.north east)$);
  \draw[fwd] (scan) -- (cache);
  \draw[fwd] (cache.east) -- (denoise.west);
  \draw[fwd] (bk.south) -- (denoise.north);
  \draw[opt] (time.west) -- (denoise.east);
  \draw[fwd] (denoise) -- (loss);

  \draw[grad] (loss.west) -| (cache.south);
  \draw[grad] ($(cache.north)+(-0.18,0)$) -- ($(scan.south)+(-0.18,0)$);
  \node[text=green!45!black, align=center] at (2.35,-1.22) {Gradient flow through\\Mamba Prefix states};

\end{tikzpicture}}
  \caption{BDLM Mamba-H training. The all-block objective applies this active-block computation at all block positions in parallel. Solid arrows show forward computation, dashed green arrows show gradient flow, and dashed black arrows mark optional timestep input. The denoiser receives a clean prefix cache, the noisy current block, and optional timestep conditioning while future blocks remain masked out. For Mamba hybrid layers, the cached prefix is forward-only and the reverse scan is local to the active block, so active-block losses train the same prefix cache object used by the sampler.}
  \label{fig:prefix-cache-objective}
\end{figure}
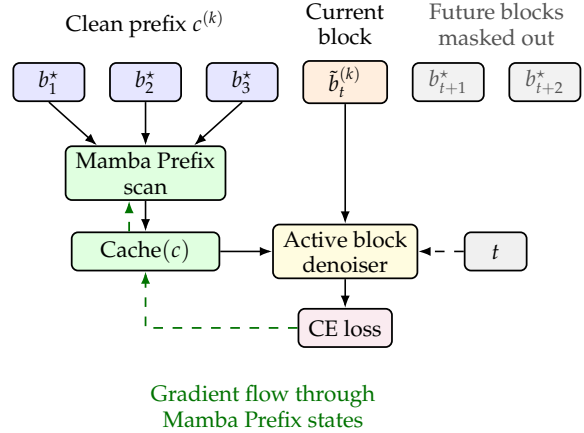
\FloatBarrier

\begin{itemize}
    \item \textbf{Cache-aligned training.} We formulate a BDLM all-block
    objective whose cached objects match inference-time block caches: attention
    KV states for attention layers and Mamba prefix states for recurrent
    layers. For the BDLM Mamba-H architecture, downstream block
    losses directly train the prefix scan that constructs the recurrent state
    reused at inference.
    \item \textbf{Controlled validation and scale-up.} We train full-sequence
    attention, full-sequence DiffuMamba-H, BDLM attention, and BDLM Mamba-H at
    87M parameters on DCLM with tuned hyperparameters. BDLM Mamba-H reaches
    the best 87M C4-en validation perplexity, 61.6 compared with 76.5 for
    BDLM attention and 83.7 and 87.6 for the full-sequence baselines. We
    scale BDLM attention and BDLM Mamba-H to 350M parameters
    with a Complete(d)P-style learning-rate transfer rule~\citep{completedp2026};
    BDLM Mamba-H remains comparable to BDLM attention in
    perplexity at this scale.
    \item \textbf{Long-context throughput gains.} We evaluate 350M training
    throughput at 8192-token context on 8x A100-80GB and 700M
    random-initialized inference throughput on 1x A100-80GB with batch size 1.
    BDLM Mamba-H overtakes BDLM attention (and
    full-sequence DiffuMamba-H in throughput) as generation length grows, reaching
    19.7x full-sequence DiffuMamba-H throughput at 65K and 3.7x BDLM
    attention throughput at 262K.
\end{itemize}

\section{Related Work}

\paragraph{Diffusion language models.}
Discrete diffusion language models corrupt text with masking or discrete noise
and train denoisers to reconstruct clean tokens
\citep{austin2021d3pm,hoogeboom2021argmax,sahoo2024mdlm,nie2025llada}.
BDLMs add an autoregressive block factorization so each completed block becomes
clean prefix context for later blocks~\citep{arriola2025blockdiffusion}.
This gives block diffusion a native cache boundary, unlike full-sequence MDLMs
whose hidden states are tied to the current noisy sequence.
Training-free cache methods retrofit reuse onto pretrained diffusion language
models through approximate KV reuse or adaptive caching
\citep{wu2025fastdllm,ma2025dkvcache,nguyentri2025attentionkv,liu2025dllmcache,wang2025discretediffusionforcing}.
Our focus is instead to train a block-diffusion architecture whose cache is part
of both the objective and the sampler.

\paragraph{Mamba hybrids for long-context generation.}
Mamba replaces attention over token pairs with a selective state-space scan
whose transition, input, and output maps are functions of the current token
\citep{gu2023mamba}. In the notation used below, a Mamba layer carries a
recurrent boundary state $s_i^\ell$ after position or block $i$ in layer
$\ell$; this state includes the convolutional state and the selective-SSM
state needed to continue the forward scan without reprocessing the prefix.
Mamba-2 recasts this family through structured state-space duality, yielding
more efficient scan kernels and a closer relationship to attention
\citep{dao2024mamba2}. Recent autoregressive systems increasingly use hybrid
stacks that preserve some full-attention layers while replacing many layers
with recurrent, linear-attention, or state-space modules: examples include
Jamba's Transformer--Mamba mixture-of-experts design, MiniMax-01's Lightning
Attention and softmax-attention mixture, Qwen3-Next's Gated DeltaNet/full-attention
hybrid, and Kimi Linear's Kimi Delta Attention/MLA hybrid
\citep{lieber2024jamba,li2025minimax01,qwen2025qwen3next,kimi2025kimilinear}.
These autoregressive models motivate hybridization as a practical long-context
scaling direction; our work studies the analogous question for block diffusion,
where cache reuse must be aligned with the denoising objective rather than with
a left-to-right next-token likelihood. DiffuMamba-H interleaves attention and bidirectional
Mamba layers for masked diffusion denoising and demonstrates that
Mamba-backed denoisers improve long-context inference throughput
\citep{singh2026diffumamba}.

\section{Background}


Mamba is a selective state-space sequence mixer that replaces attention over
tokens with a recurrent scan whose state can be continued across sequence
segments~\citep{gu2023mamba}. Mamba-2 improves this family with a structured
state-space duality and more efficient kernels~\citep{dao2024mamba2}. In
diffusion language models, DiffuMamba-H uses a sparse hybrid schedule that
interleaves attention and bidirectional Mamba layers, showing that
full-sequence diffusion denoisers can benefit from linear-time mixers
\citep{singh2026diffumamba}. The present work uses the same broad hybrid
motivation, but modifies the scan directionality so BDLM generation can reuse a
prefix cache natively at training and inference time.
At layer $\ell$, we write a simplified selective state-space update as
\begin{equation}
  s_i^\ell = A_i^\ell s_{i-1}^\ell + B_i^\ell x_i^\ell,\qquad
  y_i^\ell = C_i^\ell s_i^\ell + D^\ell x_i^\ell,
  \label{eq:mamba-scan-background}
\end{equation}
where $A_i^\ell,B_i^\ell,C_i^\ell$ are token-dependent projections and
$s_i^\ell$ is the recurrent boundary state after token or block $i$. A prefix
cache is the collection of these forward Mamba states, together with attention
KV tensors for attention layers, at a block boundary.

Masked discrete diffusion corrupts clean text by replacing tokens with masks
and trains a denoiser to recover the original tokens
\citep{austin2021structured,sahoo2024simple}. If $\tilde{x}_t$ is the
corrupted sequence at timestep $t$, a full-sequence denoiser optimizes
\begin{equation}
  \mathcal{L}_{\mathrm{full}}(\theta)
  =
  \E_{x,t,M_t}
  \left[
    \frac{1}{|M_t|}
    \sum_{i\in M_t}
    \CE(f_\theta(\tilde{x}_t,t)_i,x_i)
  \right].
  \label{eq:full}
\end{equation}
This objective gives each masked token bidirectional context over the denoising
window. It also makes the entire window part of the timestep-conditioned
computation. During generation, the hidden states are tied to the current noisy
sequence, so a stable prefix has no reusable cache boundary and must be reprocessed
at every reverse step.


BDLMs introduce that boundary by partitioning a sequence into $K$ blocks
$b^{\star(1)},\ldots,b^{\star(K)}$. Following the BDLM formulation~\citep{arriola2025blockdiffusion}, the
likelihood is autoregressive over blocks,
\begin{equation}
  \log p_\theta(x)
  =
  \sum_{k=1}^{K}
  \log p_\theta\!\left(b^{\star(k)}\,\middle|\,b^{\star(<k)}\right),
  \label{eq:bdlm-block-factorization}
\end{equation}
and each block conditional is represented by a discrete diffusion process over
only the current block~\citep{arriola2025blockdiffusion}. Applying the diffusion
NELBO to each term gives the block-summed objective
\begin{equation}
  -\log p_\theta(x)
  \le
  \mathcal{L}_{\mathrm{BD}}(x;\theta)
  :=
  \sum_{k=1}^{K}
  \mathcal{L}\!\left(b^{\star(k)},b^{\star(<k)};\theta\right).
  \label{eq:bdlm-block-nelbo}
\end{equation}
For the masked continuous-time parameterization used in BDLM-style training,
this can be written as a sum of weighted current-block denoising losses,
\begin{equation}
  \mathcal{L}_{\mathrm{BD}}(x;\theta)
  =
  \sum_{k=1}^{K}
  \E_{t,q}
  \left[
    w(t)\,
    \CE\!\left(
      f_\theta(b_t^{(k)},b^{\star(<k)},t),
      b^{\star(k)}
    \right)
  \right],
  \qquad
  w(t)=\frac{-\alpha'_t}{1-\alpha_t},
  \label{eq:bdlm-masked-objective}
\end{equation}
where $b_t^{(k)}\sim q_t(\cdot\mid b^{\star(k)})$ is the noised current block,
$b^{\star(<k)}$ is a clean prefix, and future blocks are excluded. The
all-block training objective evaluates this sum over block boundaries
within the same training window, so every block is trained as a current
denoising target conditioned on its clean prefix.

This factorization enables caching for block diffusion.
For an attention BDLM, the denoiser for block $k$ can be written with the same
cache interface used by autoregressive Transformers:
\begin{equation}
  z^{(k)}, K_k, V_k
  \leftarrow
  f_\theta^{(k)}
  \!\left(b_t^{(k)}, K_{1:k-1}, V_{1:k-1}\right)
  \equiv
  f_\theta^{(k)}
  \!\left(b_t^{(k)}, b^{\star(<k)}\right),
  \label{eq:bdlm-kv-cache}
\end{equation}
where $z^{(k)}$ are clean-token logits for the current block and
$K_{1:k-1},V_{1:k-1}$ are key/value tensors computed from completed prefix
blocks. During sampling, the model denoises all tokens within the current block
in parallel while reading this prefix cache; after the block is finalized, a
clean forward pass over the new block appends $K_k,V_k$ for subsequent blocks.
Thus cache reuse follows from the block conditional structure itself: completed
blocks are fixed inputs to later block denoisers, while reverse diffusion
updates are confined to the active block.

\paragraph{NELBO perplexity.}
For diffusion language models, exact $\log p_\theta(x)$ is generally
intractable because likelihood marginalizes over reverse denoising
trajectories. We therefore evaluate a negative evidence lower bound (NELBO).
For a validation corpus with token counts $n_j$ and sequence-level NELBO
estimates $\widehat{\mathcal{L}}_{\mathrm{NELBO}}(x_j)$, the reported
token-weighted perplexity is
\begin{equation}
  \PPL_{\mathrm{NELBO}}
  =
  \exp\!\left(
    \frac{\sum_j \widehat{\mathcal{L}}_{\mathrm{NELBO}}(x_j)}
         {\sum_j n_j}
  \right).
  \label{eq:nelbo-ppl-background}
\end{equation}
This is an upper-bound-derived perplexity metric rather than an exact
autoregressive likelihood. It lets full-sequence MDLMs and BDLMs be compared under the same sequence-token-weighted diffusion validation protocol.

\paragraph{Complete(d)P learning-rate transfer.}
For scale-up, we transfer learning rates from 87M to 350M using a
Complete(d)P-style rule~\citep{completedp2026}. The motivation is that a
larger run changes width, depth, batch size, and duration simultaneously. We use
the hidden-weight transfer form
\begin{equation}
  \eta_{\mathrm{large}}
  =
  \eta_{\mathrm{small}}\,
  r_{\mathrm{width}}^{a_w}
  r_{\mathrm{depth}}^{a_d}
  r_{\mathrm{batch}}^{a_b}
  r_{\mathrm{duration}}^{a_T},
  \label{eq:completedp-transfer}
\end{equation}
where each $r$ is the large-to-small ratio for that axis and the exponents are
the Complete(d)P transfer exponents for hidden weights. This gives scale
factors of $0.31$ for BDLM attention and $0.33$ for BDLM Mamba-H in our 87M to
350M transfer; the resulting learning rates are reported in
Table~\ref{tab:scaleup-350m}.

\section{Block Diffusion Language Model}


We use $c^{(k)}=b^{\star(<k)}$ for the clean prefix before block $k$,
$\tilde{b}^{(k)}_t$ for the noised active block at reverse timestep $t$, and
$f_\theta$ for the block denoiser. For Mamba layer $\ell$, $s_k^\ell$ denotes
the reusable forward recurrent boundary state produced by scanning the clean
prefix. The symbols $\rightarrow$ and $\leftarrow$ denote forward and reverse
Mamba scan directions, while $y_{k,t}^{\rightarrow,\ell}$ and
$y_{k,t}^{\leftarrow,\ell}$ are the corresponding active-block hidden states.

Let $\mathcal{C}^{\ell}(c)$ denote the layer-$\ell$ prefix cache. Prefix-cache
reuse requires this cache to be a function of the clean prefix alone, so $\mathcal{C}^{\ell}(c) \ \text{must be independent of}\ t$.
  
Naively reusing a diffusion denoiser can violate this requirement because
diffusion Transformers often inject timestep information through adaptive
normalization applied to every token~\citep{peebles2023scalable}. If the same
modulation touches prefix tokens, the cache becomes timestep-specific and must
be bespoke for every reverse step. We therefore factor each layer into a
timestep-free prefix operator and a timestep-conditioned active-block
continuation:
\begin{equation}
  \mathcal{C}^{\ell}(c)=P^{\ell}(c),
  \qquad
  h_{b,t}^{\ell+1}=
  B^{\ell}(h_{b,t}^{\ell};\mathcal{C}^{\ell}(c),e_t).
  \label{eq:cacheable-layer}
\end{equation}
Appendix~\ref{app:adaln} gives the corresponding AdaLN factorization.

Mamba directionality adds a second constraint. A fully bidirectional Mamba
denoiser over $[c;\tilde{b}_t]$ combines a forward scan with a reverse scan
over the reversed prefix-plus-block sequence. The reverse prefix states then
depend on the active block, so they are not prefix-only values and cannot be
reused as a BDLM prefix cache. Our BDLM Mamba hybrid instead runs the
reverse-direction Mamba only within the current denoising block, rather than
over the full sequence as in DiffuMamba (we denote this as a
\emph{block partial-reverse} architecture). BDLM Mamba caches only the forward boundary states,
\begin{equation}
  s_k^\ell = F_{\theta,\rightarrow}^{\ell}(c^{(k)}),
  \label{eq:boundary-state}
\end{equation}
and continues them into the active block:
\begin{align}
  y_{k,t}^{\rightarrow,\ell}
  &=
  F_{\theta,\rightarrow}^{\ell}(\tilde{b}_t^{(k)};s_k^\ell),\\
  y_{k,t}^{\leftarrow,\ell}
  &=
  \operatorname{rev}
  \left(
    F_{\theta,\leftarrow}^{\ell}(\operatorname{rev}(\tilde{b}_t^{(k)}))
  \right),\\
  y_{k,t}^{\ell}
  &=
  y_{k,t}^{\rightarrow,\ell}+y_{k,t}^{\leftarrow,\ell}.
  \label{eq:partial-reverse}
\end{align}
The active block still receives both left-to-right and right-to-left mixing, but
only within the current block. The reusable prefix cache remains valid because
the reverse scan is local to the active block. This serves as a native BDLM analogue of generation with block-caching: generated prefix blocks are summarized by forward recurrent state,
while denoising inside the current block keeps local bidirectionality.

The prefix-cache objective trains the same forward cached representations which will be used at inference:
\begin{equation}
  \mathcal{L}_{\mathrm{prefix}}(\theta)
  =
  \frac{1}{K}\sum_k\E_t
  \ell_\theta(b^{\star(k)}\mid s_k^{1:L},\tilde{b}_t^{(k)},t).
  \label{eq:prefix-cache}
\end{equation}
 For attention layers, the corresponding
cache object is the usual prefix key/value tensor. For Mamba layers, the cache object is the recurrent boundary state, including convolution and selective-SSM
state needed to continue the scan into the active block.

\subsection{Block Diffusion Mamba Hybrid Training}
\label{sec:bdlm-mamba-prefix-cache-training}

Consider a sequence split into blocks $b_1,b_2,b_3,b_4$, and let $b_4$ be the
active diffusion block. The model first scans the clean prefix to produce the
Mamba prefix cache
\begin{equation}
  C_4 = F_\theta(b_1,b_2,b_3),
  \label{eq:cache-c4}
\end{equation}
where $C_4$ contains the convolution and state-space states at the boundary
before $b_4$. The active block is then denoised from that same cache object:
\begin{equation}
  \ell_4 =
  \CE\!\left(D_\theta(C_4,\tilde{b}_4,t), b_4\right).
  \label{eq:prefix-cache-loss}
\end{equation}
Across all-block training, the same construction is evaluated for every block
boundary in parallel. Backpropagation follows the prefix-cache construction
path:
\begin{align}
  \ell_4 &\rightarrow C_4
  \rightarrow F_\theta(b_1,b_2,b_3)
  \rightarrow\theta .
  \label{eq:prefix-cache-gradient}
\end{align}
The cache still contains the same prefix-only Mamba states that are reused at
inference. Downstream block losses shape the representations stored in those
states, aligning the training computation with the sampler's cache interface:
the state consumed by later blocks is produced by the same prefix scan that is
used during native BDLM generation.

\section{Experiments}


\subsection{Experiment Setup}

We evaluate training and inference throughput as well as validation performance
  of our BDLM Mamba hybrid against full-sequence attention, full-sequence
  DiffuMamba-H, and BDLM attention baselines. The 87M sweep trains all four architectures on the following learning rates:  $\{5{\times}10^{-4},\,10^{-3},\,2{\times}10^{-3},\,4{\times}10^{-3},\,8{\times}10^{-3}\}$.
All configurations use 5 billion tokens sampled from DCLM~\citep{li2024datacomp},  8192-token training
  sequences, 16 diffusion steps, and no-timestep modulation; BDLM
  configurations use block size 256. The full-sequence DiffuMamba-H model uses a fully bidirectional Mamba hybrid denoiser over the
  denoising window, whereas BDLM Mamba-H only runs the reverse Mamba on the current active denoising block to leverage
  prefix-caching construction as described in
  Section~\ref{sec:bdlm-mamba-prefix-cache-training}.

Likelihood validation uses the same BDLM/MDLM-compatible diffusion-NELBO
  evaluator for all models at 8192-token context, with one Monte Carlo
  timestep/mask corruption sample per validation sequence. We report C4-en and
  Paloma-C4 NELBO perplexity and bits per byte (BPB). NELBO
  perplexity is computed as the exponential of the token-weighted average NELBO
  estimate, and BPB divides the same token-weighted negative log-likelihood in
  bits by the number of UTF-8 source-document bytes. We report MCQA separately as
  pseudo-likelihood answer-choice accuracy over fixed candidate answers.
  Masked-token accuracy from the diffusion-NELBO evaluator is included in the
  full 87M sweep in Table~\ref{tab:validation-87m-full}.

For MCQA, we evaluate HellaSwag, PIQA, ARC-Easy, ARC-Challenge, BoolQ,
  and WinoGrande
  \citep{zellers2019hellaswag,bisk2020piqa,clark2018think,clark2019boolq,sakaguchi2021winogrande}
  and report macro-average accuracy across tasks. For each example, we score each
  provided answer choice under the model's pseudo-likelihood scoring rule and
  select the highest-scoring choice. We use fixed-choice evaluation over
evaluations that rely on free-form generation and answer extraction, making it suitable for comparing
   base models with no post training . The suite includes binary-choice tasks (PIQA, BoolQ,
  and WinoGrande) and multi-choice tasks with fixed or variable answer sets
  (HellaSwag, ARC-Easy, and ARC-Challenge).

Appendix~\ref{app:timestep} compares no-timestep and timestep-conditioned
  objectives. We use the no-timestep objective in the main experiments for two
  reasons. First, it matches the original BDLM experimental convention~\citep{arriola2025blockdiffusion}. Second,
  it keeps clean-prefix caches timestep-invariant: if timestep modulation is
  applied to prefix tokens, the cached prefix states depend on the reverse
  diffusion step and cannot be reused across denoising steps. Even when a
  timestep is held fixed, a technically cache-aligned BDLM implementation should
  avoid learned timestep embeddings in the prefix-cache construction, because
  completed prefix states are properties of clean prefix text rather than of a
  particular reverse step. The AdaLN
  factorization in Appendix~\ref{app:adaln} describes how BDLM Mamba-H can make use of prefix caching; 
  timestep modulation is applied only to the active denoising block, while the
  clean-prefix scan remains timestep-free. Empirically, the fixed learning rate
  comparison in Appendix~\ref{app:timestep} shows that  timestep conditioning
  improves BDLM attention, but worsens BDLM Mamba-H and both full-sequence baselines
  on NELBO perplexity. We therefore use the no-timestep objective for the main
  model-selection and scale-up experiments.

  For the 350M scale-up experiments, we transfer learning rates from the selected 87M configurations
  using a Complete(d)P-style hidden-weight rule~\citep{completedp2026}. We use
  this rule as a principled way to choose scale-up candidates because the 350M
  runs change width, depth, global batch size, and training horizon
  simultaneously. Recent diffusion-language-model
  scaling studies that find learning rate and batch size transfer rules remain
  useful for DLMs~\citep{ni2025quokka}, and with work using CompleteP for stable
  learning-rate transfer across width and depth in discrete DLM scaling
  \citep{vonrutte2026scaling}. The resulting learning rates are shown in
  Table~\ref{tab:scaleup-350m}. We select the best 87M models by C4-en heldout perplexity: $4{\times}10^{-3}$ for BDLM attention
  and $8{\times}10^{-3}$ for BDLM Mamba-H. Both 350M models use the same warmup length, global batch, sequence length,
  data stream, and no-timestep objective. We choose model-specific optimizer-step counts so each run trains just above the same target token-to-parameter ratio, then vary only the architecture and Complete(d)P-transferred learning rate.

\subsection{Language Modeling Quality}

We select one checkpoint per architecture by the best C4-en
  $\mathrm{PPL}_{\mathrm{NELBO}}$ in the 87M-class learning-rate sweep. Under this
  selection rule, BDLM Mamba-H gives the lowest likelihood-bound validation
  perplexity: 61.6, compared with 76.5 for BDLM attention, 83.7 for
  full-sequence DiffuMamba-H, and 87.6 for full-sequence attention; the full
  20-row sweep is present in Appendix Table~\ref{tab:validation-87m-full}. MCQA provides a complementary fixed-choice
  preference evaluation separate to the NELBO-perplexity
  ranking.

  At 350M scale, the two selected BDLM models remain comparable with BDLM attention obtaining
  37.0 C4-en $\mathrm{PPL}_{\mathrm{NELBO}}$ and 36.1 Paloma-C4
  $\mathrm{PPL}_{\mathrm{NELBO}}$, while BDLM Mamba-H obtains 38.2 and 36.8,
  respectively. The corresponding MCQA accuracies are 41.3 for BDLM attention
  and 43.3 for BDLM Mamba-H. Thus we show BDLM Mamba-H remains in the
  same validation-quality range while enabling native long-context cached
  inference.


\begin{table}[t]
  \centering
  \caption{Best 87M-class DCLM~\citep{li2024datacomp} no-timestep validation
  configurations, selected by C4-en NELBO perplexity. PPL denotes the BDLM/MDLM-compatible
  diffusion-NELBO validation path with one timestep sample at 8192-token context.
  All models are trained on 5.00B DCLM tokens. BPB divides token-weighted negative
  log-likelihood in bits by the number of UTF-8 bytes in the source documents.
  MCQA is pseudo-likelihood accuracy over fixed answer choices. The full
  learning-rate sweep is reported in Table~\ref{tab:validation-87m-full}.}
  \label{tab:validation-87m-best}
  \scriptsize
  \setlength{\tabcolsep}{3.0pt}
  \resizebox{\columnwidth}{!}{%
  \begin{tabular}{llrrrrr}
    \toprule
    Model & LR & C4 PPL & C4 BPB & Paloma C4 PPL & Paloma BPB & MCQA \\
    \midrule
    Partially Reverse BDLM Mamba-H & $8{\times}10^{-3}$ & \textbf{61.59} & \textbf{1.30} & \textbf{58.48} & \textbf{1.28} & 35.58 \\
    BDLM attention & $4{\times}10^{-3}$ & 76.46 & 1.37 & 70.81 & 1.34 & 36.31 \\
    Full-sequence DiffuMamba-H & $4{\times}10^{-3}$ & 83.73 & 1.39 & 80.42 & 1.38 & 37.32 \\
    Full-sequence attention & $8{\times}10^{-3}$ & 87.64 & 1.41 & 83.25 & 1.39 & 35.03 \\
    \bottomrule
  \end{tabular}}
\end{table}

\begin{table}[t]
  \centering
\caption{350M BDLM scale-up configuration. Learning rates are transferred from
  the best 87M C4-en validation settings using a Complete(d)P-style hidden-weight
  transfer rule~\citep{completedp2026}. Both models use no timestep conditioning, DCLM, the BDLM~\citep{arriola2025blockdiffusion}
  objective, 8192-token training sequences, global batch 64, 2000 warmup steps,
  the same optimizer-step horizon, and 17.65B DCLM training tokens.}
  \label{tab:scaleup-350m}
  \scriptsize
  \setlength{\tabcolsep}{3.0pt}
  \resizebox{\columnwidth}{!}{%
  \begin{tabular}{lrrrrrr}
    \toprule
    Model & Params & 87M LR & Transfer factor & 350M LR & Steps & Token:param \\
    \midrule
    BDLM attention & 353.03M & $4{\times}10^{-3}$ & 0.31 & $1.23{\times}10^{-3}$ & 33{,}668 & 50.00 \\
    Partially Reverse BDLM Mamba-H & 350.97M & $8{\times}10^{-3}$ & 0.33 & $2.65{\times}10^{-3}$ & 33{,}668 & 50.29 \\
    \bottomrule
  \end{tabular}}
\end{table}

\begin{table}[t]
  \centering
  \caption{350M BDLM no-timestep validation results. C4 and Paloma C4 are
  evaluated at 8192-token context with 256 examples per dataset. BPB divides
  token-weighted negative log-likelihood in bits by the number of UTF-8 bytes in
  the source documents. MCQA is pseudo-likelihood accuracy over fixed answer
  choices.}
  \label{tab:validation-350m}
  \scriptsize
  \setlength{\tabcolsep}{3.0pt}
  \resizebox{\columnwidth}{!}{%
  \begin{tabular}{lrrrrrr}
    \toprule
    Model & LR & C4 PPL & C4 BPB & Paloma C4 PPL & Paloma BPB & MCQA \\
    \midrule
    BDLM attention & $1.23{\times}10^{-3}$ & \textbf{36.97} & \textbf{1.14} & \textbf{36.14} & \textbf{1.13} & 41.28 \\
    Partially Reverse BDLM Mamba-H & $2.65{\times}10^{-3}$ & 38.24 & 1.15 & 36.81 & 1.13 & \textbf{43.29} \\
    \bottomrule
  \end{tabular}}
\end{table}

\FloatBarrier

\subsection{Training and Inference Throughput}

  Training throughput is
  measured at 350M scale on 8x A100-80GB with synthetic 8192-token batches.
  Inference throughput is measured with 700M random-initialized models on
  1x A100-80GB with BF16 and batch size 1.  Both benchmarks use
  5 warmup iterations and average over 15 measured iterations. The context generation lengths
  follow the DiffuMamba long-context protocol: full-sequence denoisers are
  measured through the 65K-token regime, while native BDLM models are
  extended along the block-cached ladder up to 262k generated tokens.

  At 350M scale, all four architectures use the target global batch on
  8x A100-80GB and fit into memory without activation checkpointing. The full-sequence models are
  significantly faster in 8192-token training throughput, reflecting the additional
  cost of evaluating the all-block BDLM objective during training. Among the BDLM
  models, BDLM Mamba-H trains faster than BDLM attention and uses less peak
  reserved memory. At inference time, the
  native BDLM models retain substantially higher throughput than full-sequence
  denoisers as generation length grows. BDLM attention is faster at short lengths
  in our implementation, while BDLM Mamba-H overtakes it
  in the long-context regime starting at approximately 16k generated tokens.

\setcounter{table}{4}
\begin{table}[!htbp]
  \centering
  \caption{350M 8x A100-80GB training throughput at 8192-token context. All
  models use synthetic batches, global batch 64, BF16, ZeRO-2, tiled linear
  cross-entropy, and FlashAttention kernels when available.}
  \label{tab:training-throughput-8k}
  \scriptsize
  \setlength{\tabcolsep}{3.0pt}
  \resizebox{0.76\columnwidth}{!}{%
  \begin{tabular}{lrrr}
    \toprule
    Model & GPUs & Global batch & tok/s \\
    \midrule
    Full-sequence attention & 8x A100 & 64 & 229{,}957 \\
    Full-sequence DiffuMamba-H & 8x A100 & 64 & 223{,}825 \\
    BDLM attention & 8x A100 & 64 & 151{,}976 \\
    Partially Reverse BDLM Mamba-H & 8x A100 & 64 & 166{,}849 \\
    \bottomrule
  \end{tabular}}
\end{table}
\FloatBarrier

  At 65K, BDLM Mamba-H is 19.7x faster than the full-sequence
  DiffuMamba-H baseline at the same length. On the extended BDLM ladder, BDLM Mamba-H is 3.7x faster than
  BDLM attention at 262K. This crossover is qualitatively consistent with
  DiffuMamba's finding that Mamba-backed block generation increasingly separates
  from attention-backed block generation as sequence length
  grows~\citep{singh2026diffumamba}. Raw tokens/s, implementation details, and
  length-by-length speedup ratios are given in
  Appendix~\ref{app:inference-throughput}.

\begin{figure}[t]
  \centering
  \begin{minipage}[t]{0.49\columnwidth}
    \centering
    \includegraphics[width=\linewidth]{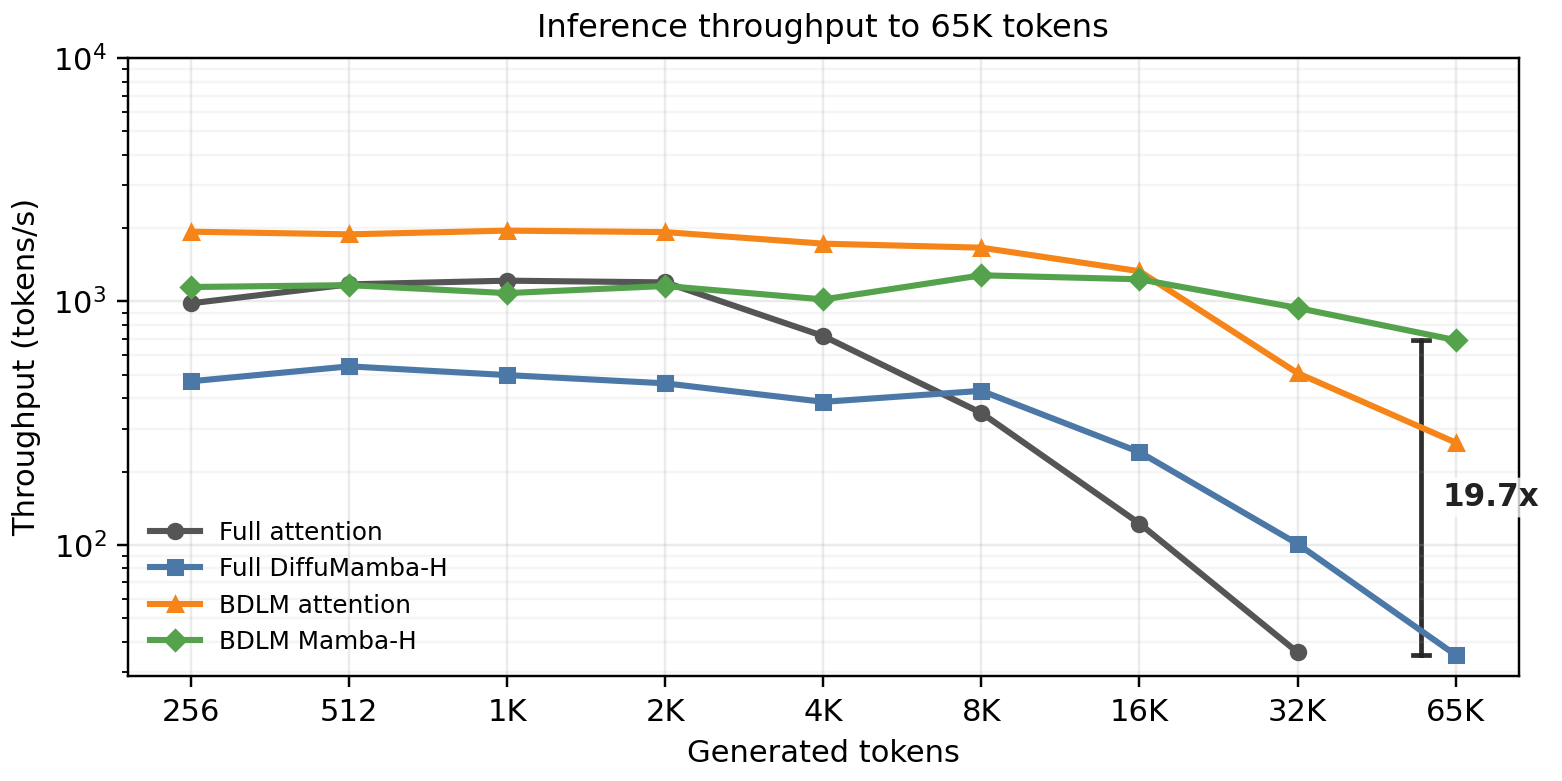}
    \vspace{-0.8em}
    \centerline{\small (a) Throughput to 65K tokens.}
  \end{minipage}
  \hfill
  \begin{minipage}[t]{0.49\columnwidth}
    \centering
    \includegraphics[width=\linewidth]{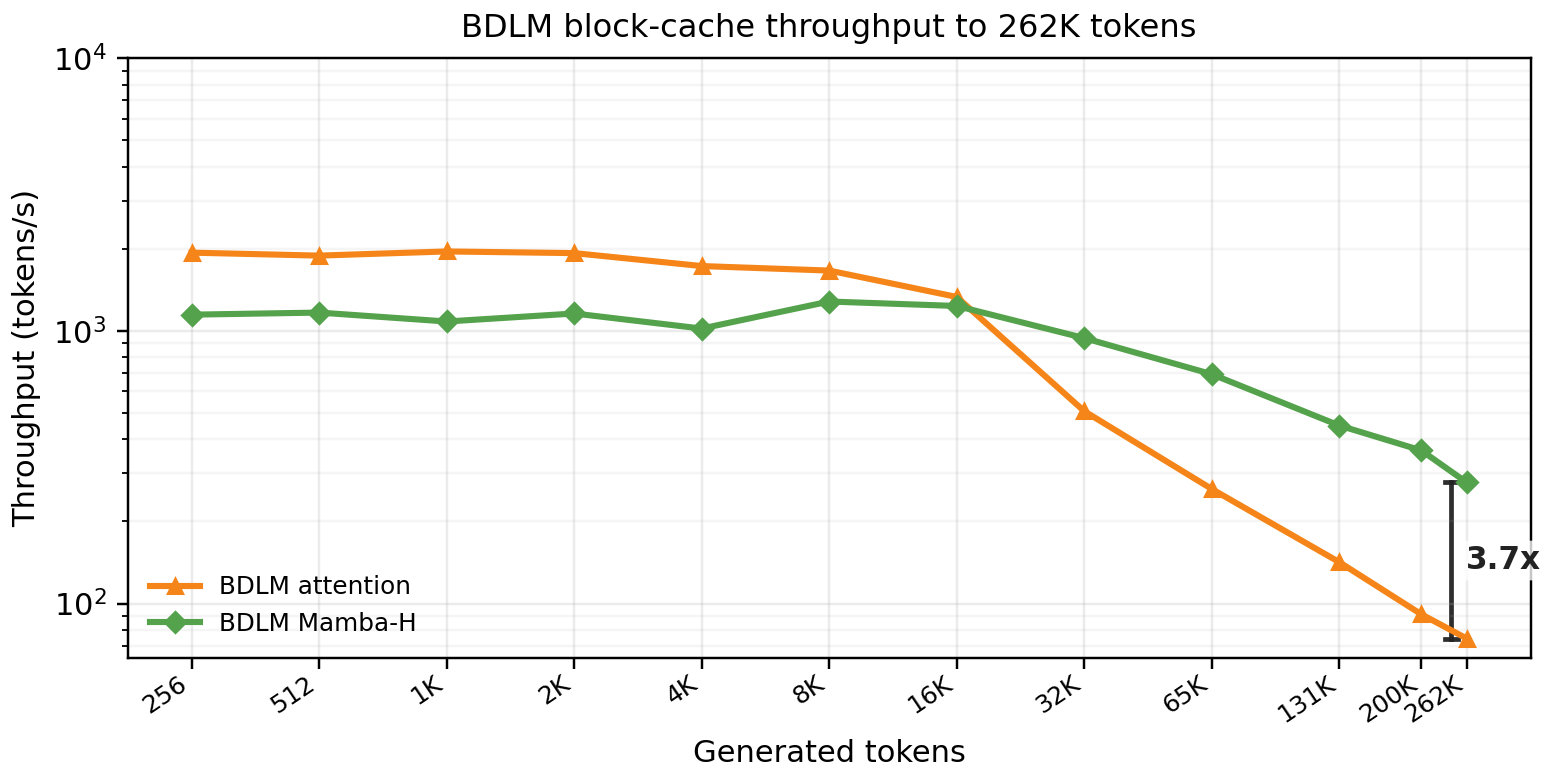}
    \vspace{-0.8em}
    \centerline{\small (b) BDLM throughput on the extended ladder.}
  \end{minipage}
  \caption{700M random-initialized single-A100 inference throughput.
  Full-sequence baselines use repeated full-sequence denoising with CUDA graphs and
  are reported. Native BDLM models use
  block size 256, BF16, cacheable prefix states, and optimized cached denoising;
  the extended ladder includes completed BDLM attention and BDLM
  Mamba-H points through 262K generated tokens.}
  \label{fig:inference-throughput-350m}
  \label{fig:inference-throughput-350m-262k}
\end{figure}

\section{Limitations}

While the block objective improves long-context
generation, we find that at 350M parameters on 8x A100-80GB, BDLM attention trains 33.9\% slower than
the full-sequence attention baseline, and BDLM Mamba-H trains 25.5\% slower than the
full-sequence DiffuMamba-H baseline. We leave it to future work to investigate optimal scaling laws for hyperparameter transfer for dLLMs and hybrid variants thereof. Complete(d)P provided a parameterization-based method of transferring learning rates across width, depth,
batch, and duration~\citep{completedp2026}, but did not explicitly derive these laws for dLLMs. Masked and block diffusion objectives have different timestep sampling,
loss weighting, and caching constraints, so the transferred 350M
learning rates are strong candidates rather than a proof of optimality.

\section{Conclusion}

We presented a training recipe for block diffusion Mamba hybrids that trains the
same prefix cache used during native BDLM generation. The method keeps the
prefix scan timestep-free to optionally support valid AdaLN timestep
conditioning for BDLM Mamba, confines reverse-direction Mamba mixing to the
active block rather than applying a full-window reverse pass, and trains the
prefix cache through downstream block losses.

Our trained models preserve validation quality while improving the
long-context systems regime. At 87M parameters, BDLM Mamba-H reaches the best
C4-en PPL in the DCLM sweep, and at 350M parameters its PPL remains in the same
range as BDLM attention. In
inference, the native BDLM models extend to lengths that full-sequence denoisers
do not practically cover, and BDLM Mamba-H increasingly separates at long
context, reaching 19.7x the full-sequence DiffuMamba-H counterpart at 65K
tokens and 3.7x BDLM attention throughput at 262K. We show that training block partial-reverse BDLM Mamba hybrids yields a
practical long-context diffusion architecture worth further exploring.

\section*{Acknowledgements}

We gratefully acknowledge Modal Labs for generously supporting this work with a Modal for Academics compute grant.

\bibliographystyle{colm2026_conference}
\bibliography{references}

\clearpage
\appendix

\section{Full 87M Validation Sweep}
\label{app:validation-sweep}

\setcounter{table}{3}
\begin{table}[!htbp]
  \centering
  \caption{Full 87M DCLM no-timestep validation sweep. All
  checkpoints are evaluated at step 19{,}074 with the same 8192-token
  diffusion-NELBO validation path used in Table~\ref{tab:validation-87m-best}.}
  \label{tab:validation-87m-full}
  \scriptsize
  \setlength{\tabcolsep}{2.2pt}
  \resizebox{\textwidth}{!}{%
  \begin{tabular}{llrrrrrrr}
    \toprule
    Architecture & LR & C4 PPL & C4 BPB & C4 acc. & Paloma PPL & Paloma BPB & Paloma acc. & MCQA \\
    \midrule
    Full attention & $5{\times}10^{-4}$ & 111.59 & 1.48 & 10.94 & 106.39 & 1.47 & 11.32 & 33.76 \\
    Full attention & $10^{-3}$ & 101.48 & 1.46 & 11.47 & 95.39 & 1.43 & 11.78 & 37.31 \\
    Full attention & $2{\times}10^{-3}$ & 98.78 & 1.45 & 11.45 & 94.12 & 1.43 & 11.70 & 36.79 \\
    Full attention & $4{\times}10^{-3}$ & 89.02 & 1.41 & 11.98 & 83.80 & 1.39 & 12.41 & 33.95 \\
    Full attention & $8{\times}10^{-3}$ & 87.64 & 1.41 & 12.04 & 83.25 & 1.39 & 12.43 & 35.03 \\
    \midrule
    Full-sequence DiffuMamba-H & $5{\times}10^{-4}$ & 104.39 & 1.46 & 11.27 & 98.98 & 1.44 & 11.55 & 36.04 \\
    Full-sequence DiffuMamba-H & $10^{-3}$ & 95.52 & 1.44 & 11.50 & 92.33 & 1.42 & 11.82 & 35.52 \\
    Full-sequence DiffuMamba-H & $2{\times}10^{-3}$ & 93.97 & 1.43 & 11.53 & 90.31 & 1.42 & 11.95 & 35.67 \\
    Full-sequence DiffuMamba-H & $4{\times}10^{-3}$ & 83.73 & 1.39 & 12.08 & 80.42 & 1.38 & 12.50 & 37.32 \\
    Full-sequence DiffuMamba-H & $8{\times}10^{-3}$ & 85.50 & 1.40 & 12.00 & 81.46 & 1.38 & 12.41 & 34.49 \\
    \midrule
    BDLM attention & $5{\times}10^{-4}$ & 91.35 & 1.42 & 10.60 & 84.46 & 1.40 & 10.72 & 41.04 \\
    BDLM attention & $10^{-3}$ & 83.97 & 1.40 & 10.95 & 76.86 & 1.37 & 11.10 & 40.08 \\
    BDLM attention & $2{\times}10^{-3}$ & 77.10 & 1.37 & 11.24 & 71.05 & 1.34 & 11.39 & 41.50 \\
    BDLM attention & $4{\times}10^{-3}$ & 76.46 & 1.37 & 11.29 & 70.81 & 1.34 & 11.43 & 36.31 \\
    BDLM attention & $8{\times}10^{-3}$ & 80.43 & 1.38 & 11.09 & 73.54 & 1.35 & 11.24 & 36.56 \\
    \midrule
    Partially Reverse BDLM Mamba-H & $5{\times}10^{-4}$ & 81.94 & 1.39 & 10.80 & 74.46 & 1.36 & 11.02 & 34.76 \\
    Partially Reverse BDLM Mamba-H & $10^{-3}$ & 72.97 & 1.35 & 11.17 & 67.06 & 1.32 & 11.41 & 36.48 \\
    Partially Reverse BDLM Mamba-H & $2{\times}10^{-3}$ & 67.02 & 1.32 & 11.60 & 62.69 & 1.30 & 11.66 & 34.21 \\
    Partially Reverse BDLM Mamba-H & $4{\times}10^{-3}$ & 62.08 & 1.30 & 11.90 & 58.42 & 1.28 & 12.00 & 35.82 \\
    Partially Reverse BDLM Mamba-H & $8{\times}10^{-3}$ & 61.59 & 1.30 & 11.92 & 58.48 & 1.28 & 11.99 & 35.58 \\
    \bottomrule
  \end{tabular}}
\end{table}
\setcounter{table}{5}

\FloatBarrier

\section{Inference Throughput Protocol and Raw Results}
\label{app:inference-throughput}

All inference models use the same block size, denoising-step ratio, precision,
and measurement policy. Full-sequence baselines use fixed-shape CUDA graphs for
repeated full-window denoising and FlashAttention kernels where available
\citep{dao2022flashattention}. BDLM attention uses preallocated mutable prefix
KV caches, tiled key-cache append, BDLM-style FlashAttention for prefix-cache
and active-block attention, graph-captured block denoising steps, and tiled
LM-head evaluation. BDLM Mamba-H uses the block partial-reverse Mamba construction:
native Mamba kernels, chunked prefix prefill, read-only recurrent cache
assumptions, compiled fixed-shape Mamba block continuations with CUDA graphs
enabled, graph-captured block denoising steps, and the same chunked LM-head
path.

Table~\ref{tab:inference-throughput} gives the raw inference throughput numbers
used in Figure~\ref{fig:inference-throughput-350m}. In our implementation,
BDLM attention is faster at short lengths, where the KV cache is still small
and softmax attention remains efficient. BDLM Mamba-H overtakes BDLM attention at
longer lengths: 1.9x at 32K, 2.6x at 65K, 3.2x at 131K, 4.0x at 200K, and
3.7x at 262K relative to BDLM attention. These
absolute ratios are specific to the 700M random-initialized A100 benchmark used
here.

\begin{table}[h]
  \centering
  \caption{Raw 700M random-initialized inference throughput in tokens/s. Values
  are measured on 1x A100-80GB with BF16 and batch size 1. Full-sequence baselines
  use CUDA graphs for fixed-shape denoising. Native BDLM models use block size 256
  and optimized cached denoising with graph-captured block steps where
  supported.}
  \label{tab:inference-throughput}
  \scriptsize
  \setlength{\tabcolsep}{2.4pt}
  \resizebox{\textwidth}{!}{%
  \begin{tabular}{lrrrrrrrrrrrr}
    \toprule
    Model & 256 & 512 & 1K & 2K & 4K & 8K & 16K & 32K & 65K & 131K & 200K & 262K \\
    \midrule
    Full-sequence attention & 1{,}035 & 915 & 1{,}421 & 1{,}062 & 761 & 351 & 122 & 36 & -- & -- & -- & -- \\
    Full-sequence DiffuMamba-H & 470 & 541 & 498 & 461 & 387 & 429 & 240 & 101 & 35 & -- & -- & -- \\
    BDLM attention & 1{,}935 & 1{,}887 & 1{,}955 & 1{,}927 & 1{,}728 & 1{,}663 & 1{,}331 & 508 & 263 & 142 & 91 & 74 \\
    Partially Reverse BDLM Mamba-H & 1{,}146 & 1{,}167 & 1{,}082 & 1{,}158 & 1{,}019 & 1{,}281 & 1{,}234 & 941 & 694 & 450 & 364 & 278 \\
    \bottomrule
  \end{tabular}}
\end{table}

\section{Model Architecture Details}
\label{app:architecture}

Tables~\ref{tab:arch-87m} and~\ref{tab:arch-350m} give the model geometry used
for the reported 87M sweep and 350M scale-up configurations. All configurations use an MLP ratio of
$4$, timestep embedding dimension $256$, Mamba convolution width $4$, and Mamba
expansion factor $2$. DiffuMamba-H hybrid configurations use attention at layers
$0,6,12,\ldots$ and Mamba-2 mixers in the intervening layers. The BDLM Mamba-H
architecture keeps that sparse-attention hybrid schedule, but uses the block partial-reverse
 construction described in Section~\ref{sec:bdlm-mamba-prefix-cache-training}
rather than a fully bidirectional full-sequence denoiser.

\begin{table}[h]
  \centering
  \caption{87M architecture details. Parameter counts are total trainable
  parameters for the released model definitions. Hybrid models use attention at
  layers $0$ and $6$.}
  \label{tab:arch-87m}
  \scriptsize
  \setlength{\tabcolsep}{2.6pt}
  \resizebox{\columnwidth}{!}{%
  \begin{tabular}{lrrrrrl}
    \toprule
    Model & Width & Layers & Heads & Head dim. & Mamba state/chunk & Params \\
    \midrule
    Full-sequence attention & 448 & 12 & 8 & 56 & -- & 89.17M \\
    Full-sequence DiffuMamba-H & 448 & 12 & 8 & 56 & 64 / 128 & 96.98M \\
    BDLM attention & 448 & 12 & 8 & 56 & -- & 89.17M \\
    BDLM Mamba-H & 448 & 12 & 8 & 56 & 64 / 128 & 96.98M \\
    \bottomrule
  \end{tabular}}
\end{table}

\begin{table}[h]
  \centering
  \caption{350M architecture details. Parameter counts are total trainable
  parameters for the released model definitions. Hybrid models use attention at
  layers $0$, $6$, and $12$; the BDLM Mamba-H architecture uses a width-adjusted
  350M-class geometry to match the BDLM attention scale.}
  \label{tab:arch-350m}
  \scriptsize
  \setlength{\tabcolsep}{2.6pt}
  \resizebox{\columnwidth}{!}{%
  \begin{tabular}{lrrrrrl}
    \toprule
    Model & Width & Layers & Heads & Head dim. & Mamba state/chunk & Params \\
    \midrule
    Full-sequence attention & 896 & 18 & 14 & 64 & -- & 353.03M \\
    Full-sequence DiffuMamba-H & 896 & 18 & 14 & 64 & 128 / 128 & 399.53M \\
    BDLM attention & 896 & 18 & 14 & 64 & -- & 353.03M \\
    BDLM Mamba-H & 832 & 18 & 16 & 52 & 128 / 128 & 350.97M \\
    \bottomrule
  \end{tabular}}
\end{table}

\begin{table}[h]
  \centering
  \caption{Key training and inference hyperparameters used across the reported
  DCLM experiments unless stated otherwise. Learning-rate values differ across
  the 87M sweep and 350M scale-up configurations as shown in
  Tables~\ref{tab:validation-87m-best} and~\ref{tab:scaleup-350m}.}
  \label{tab:key-hyperparameters}
  \tiny
  \setlength{\tabcolsep}{2.0pt}
  \renewcommand{\arraystretch}{0.95}
  \resizebox{\columnwidth}{!}{%
  \begin{tabular}{llll}
    \toprule
    Category & Setting & Category & Setting \\
    \midrule
    Backbones & Attention, DiffuMamba-H, BDLM attention, BDLM Mamba-H
      & Diffusion type & Absorbing-state masked diffusion \\
    Training objective & BDLM/MDLM continuous-time diffusion-NELBO objective~\citep{arriola2025blockdiffusion}
      & Parameterization & Substitution denoising target \\
    Noise schedule & Log-linear; $t \sim \mathrm{Uniform}(10^{-3}, 1)$
      & Timestep conditioning & Disabled for main no-timestep sweep and 350M scale-up \\
    Dataset & DCLM baseline, streaming document packing
      & Tokenizer & GPT-2 BPE, vocabulary size 50{,}257 plus mask token \\
    Training sequence length & 8192 tokens
      & RoPE context capacity & 262{,}144 tokens, $\theta=10{,}000$ \\
    BDLM block length & 256 tokens
      & Diffusion steps / factor $p$ & 16 \\
    Precision & BF16
      & Distributed training & DeepSpeed ZeRO-2 \\
    Activation checkpointing & Disabled
      & Optimizer & AdamW \\
    Adam betas / epsilon & $(0.9, 0.95)$ / $10^{-8}$
      & Weight decay & 0.10 \\
    Gradient clipping & 1.00
      & LR scheduler & Linear warmup, then cosine decay to $10^{-6}$ \\
    Warmup steps & 2000
      & EMA & Not used \\
    Antithetic sampling & Not used
      & 87M global batch / steps & 32 / 19{,}074 \\
    87M LR sweep & $5{\times}10^{-4}$, $10^{-3}$, $2{\times}10^{-3}$, $4{\times}10^{-3}$, $8{\times}10^{-3}$
      & 350M global batch / steps & 64 / 33{,}668 \\
    350M max learning rates & $1.23{\times}10^{-3}$ for BDLM attention; $2.65{\times}10^{-3}$ for BDLM Mamba-H
      & Inference batch size & 1 \\
    Inference precision & BF16
      & Inference block length & 256 tokens for native BDLM models \\
    Inference denoising factors & $p \in \{8,16\}$ in sweeps; reported settings use $p=16$
      & & \\
    \bottomrule
  \end{tabular}}
  \renewcommand{\arraystretch}{1.0}
\end{table}

\FloatBarrier

\section{No-Timestep Versus Timestep Conditioning}
\label{app:timestep}

Table~\ref{tab:timestep-appendix} compares no-timestep and original timestep conditioning at the 87M fixed-learning-rate slice used during recipe validation. The relevant question for this paper is whether BDLM Mamba-H can be trained with a timestep-free prefix path without losing the validation signal needed for model selection.

\begin{table}[h]
  \centering
  \caption{87M DCLM conditioning comparison at LR $4{\times}10^{-3}$ and 8192-token validation context. PPL is the BDLM/MDLM-compatible diffusion-NELBO perplexity. No-timestep values are copied from the main 87M sweep; timestep-conditioned models use the DDiT-normalization path and the same one-sample BDLM/MDLM-compatible validation protocol.}
  \label{tab:timestep-appendix}
  \scriptsize
  \setlength{\tabcolsep}{2.6pt}
  \resizebox{\columnwidth}{!}{%
  \begin{tabular}{llrrrr}
    \toprule
    Model & Objective & No-timestep C4 PPL & Timestep C4 PPL & No-timestep MCQA & Timestep MCQA \\
    \midrule
    Full attention & MDLM & 89.02 & 92.22 & 33.95 & 41.92 \\
    Full DiffuMamba-H & MDLM & 83.73 & 89.58 & 37.32 & 42.51 \\
    BDLM attention & BDLM & 76.46 & 64.13 & 36.31 & 44.85 \\
    BDLM Mamba-H & BDLM & 62.08 & 65.55 & 35.82 & 40.70 \\
    \bottomrule
  \end{tabular}}
\end{table}

\FloatBarrier

We find timestep conditioning improves BDLM attention at this learning rate, slightly worsens BDLM Mamba-H, and worsens both full-sequence baselines. We therefore use the five-learning-rate no-timestep sweep for main model selection, matching the original BDLM experimental convention~\citep{arriola2025blockdiffusion} and preserving a timestep-invariant prefix path. Timestep conditioning remains architecturally supported through the active-block-only AdaLN factorization in Appendix~\ref{app:adaln}, which keeps clean-prefix caches independent of the reverse timestep.

\section{AdaLN Factorization}
\label{app:adaln}

The architectural requirement for reusable prefix states is that they do not depend on the reverse timestep. Standard diffusion Transformers often inject timestep conditioning into every layer through adaptive normalization. If that modulation touches prefix tokens, then the layer cache becomes
\begin{equation}
  \mathcal{C}^{\ell}(c,t)
  =
  F_{\mathrm{prefix}}^{\ell}(c,e_t),
  \label{eq:timestep-cache-bad}
\end{equation}
which changes at every reverse step and cannot be reused across the active block's denoising trajectory. Adaptive layer normalization commonly maps a timestep embedding $e_t$ into shift, scale, and gate parameters and applies them before mixer and feed-forward updates:
\begin{align}
  (\delta_1^\ell,\gamma_1^\ell,g_1^\ell,\delta_2^\ell,\gamma_2^\ell,g_2^\ell)
  &=
  A^\ell(e_t),\\
  u_i^\ell(t)
  &=
  (1+\gamma_1^\ell)\odot\operatorname{LN}(h_i^\ell)+\delta_1^\ell.
\end{align}
For a full-sequence denoiser this modulation can touch every token. For a BDLM prefix cache, applying it to prefix tokens would make cached states timestep-dependent and force a different cache for every reverse step. The cacheable layer therefore factors as
\begin{equation}
  \mathcal{C}^{\ell}(c)=F_{\mathrm{prefix}}^{\ell}(c),
  \qquad
  h_{b,t}^{\ell+1}
  =
  F_{\mathrm{active}}^{\ell}
  \left(h_{b,t}^{\ell},\mathcal{C}^{\ell}(c),e_t\right).
  \label{eq:cacheable-adaln}
\end{equation}
The clean prefix path is evaluated once without timestep modulation; AdaLN and timestep gating are applied only to the active-block continuation. This remains the cleaner cache semantics even if an implementation uses a fixed timestep for all prefix tokens: the cached state should summarize completed text, not a learned timestep embedding attached to that text.

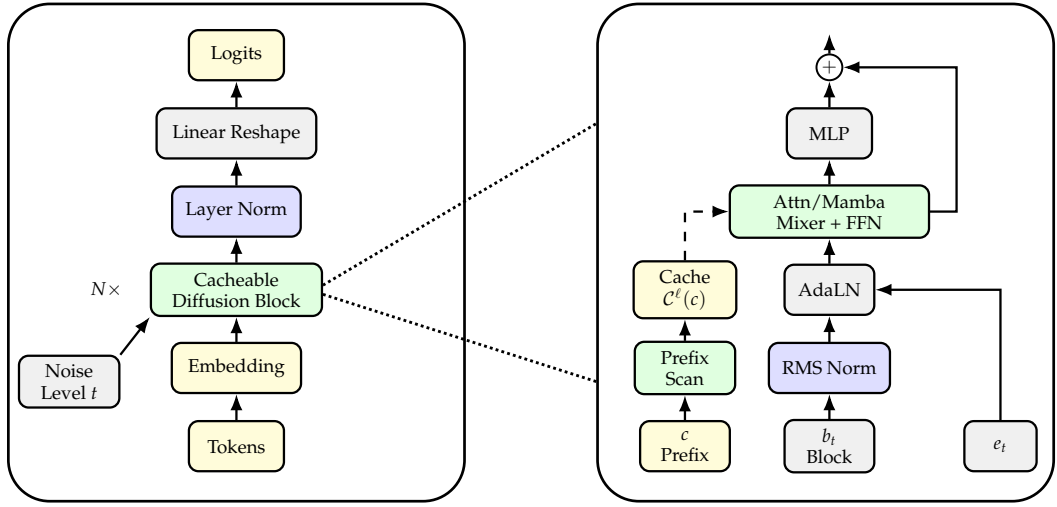
\begin{figure}[t]
  \centering
  \resizebox{\columnwidth}{!}{\begin{tikzpicture}[
  scale=0.84,
  transform shape,
  font=\scriptsize,
  box/.style={draw=black, line width=0.72pt, rounded corners=3pt, align=center, inner xsep=5pt, inner ysep=3.2pt, minimum height=0.66cm},
  greenbox/.style={box, fill=green!12},
  yellowbox/.style={box, fill=yellow!18},
  accentbox/.style={box, fill=gray!12},
  bluebox/.style={box, fill=blue!13},
  arr/.style={-{Latex[length=1.8mm]}, line width=0.72pt},
  dashedlink/.style={densely dotted, line width=0.90pt},
  reuse/.style={-{Latex[length=1.8mm]}, line width=0.72pt, dashed}
]
  \coordinate (leftframe) at (0,0);
  \coordinate (rightframe) at (7.75,0);

  \node[yellowbox, minimum width=1.25cm] (tokens) at ($(leftframe)+(0,-2.55)$) {Tokens};
  \node[yellowbox, minimum width=1.70cm] (emb) at ($(leftframe)+(0,-1.52)$) {Embedding};
  \node[accentbox, minimum width=1.32cm] (noise) at ($(leftframe)+(-2.20,-1.70)$) {Noise\\Level $t$};
  \node[greenbox, minimum width=2.25cm] (block) at ($(leftframe)+(0,-0.49)$) {Cacheable\\Diffusion Block};
  \node[bluebox, minimum width=1.65cm] (ln) at ($(leftframe)+(0,0.54)$) {Layer Norm};
  \node[accentbox, minimum width=2.10cm] (reshape) at ($(leftframe)+(0,1.57)$) {Linear Reshape};
  \node[yellowbox, minimum width=1.25cm] (logits) at ($(leftframe)+(0,2.60)$) {Logits};
  \node[left=0.24cm of block] (nblocks) {$N\times$};

  \draw[arr] (tokens) -- (emb);
  \draw[arr] (emb) -- (block);
  \draw[arr] (noise.north east) -- (block.south west);
  \draw[arr] (block) -- (ln);
  \draw[arr] (ln) -- (reshape);
  \draw[arr] (reshape) -- (logits);

  \node[yellowbox, minimum width=1.18cm] (prefix) at ($(rightframe)+(-1.85,-2.55)$) {$c$\\Prefix};
  \node[accentbox, minimum width=1.18cm] (active) at ($(rightframe)+(0.05,-2.55)$) {$b_t$\\Block};
  \node[accentbox, minimum width=0.95cm] (time) at ($(rightframe)+(2.30,-2.55)$) {$e_t$};
  \node[greenbox, minimum width=1.32cm] (prefixscan) at ($(rightframe)+(-1.85,-1.52)$) {Prefix\\Scan};
  \node[yellowbox, minimum width=1.35cm] (cache) at ($(rightframe)+(-1.85,-0.49)$) {Cache\\$\mathcal{C}^{\ell}(c)$};
  \node[bluebox, minimum width=1.46cm] (rms) at ($(rightframe)+(0.05,-1.52)$) {RMS Norm};
  \node[accentbox, minimum width=1.18cm] (adaln) at ($(rightframe)+(0.05,-0.49)$) {AdaLN};
  \node[greenbox, minimum width=2.62cm] (mixer) at ($(rightframe)+(0.05,0.54)$) {Attn/Mamba\\Mixer + FFN};
  \node[accentbox, minimum width=1.12cm] (mlp) at ($(rightframe)+(0.05,1.57)$) {MLP};
  \node[circle, draw=black, line width=0.62pt, minimum size=0.34cm, inner sep=0pt] (plus) at ($(rightframe)+(0.05,2.44)$) {$+$};

  \draw[arr] (prefix) -- (prefixscan);
  \draw[arr] (prefixscan) -- (cache);
  \draw[arr] (active) -- (rms);
  \draw[arr] (rms) -- (adaln);
  \draw[arr] (adaln) -- (mixer);
  \draw[arr] (mixer.north) -- (mlp.south);
  \draw[arr] (mixer.east) -- ++(0.35,0) |- (plus.east);
  \draw[arr] (mlp.north) -- (plus.south);
  \draw[arr] (plus.north) -- ++(0,0.26);
  \draw[reuse] (cache.north) -- (cache.north |- mixer.west) -- (mixer.west);
  \draw[arr] (time.north) -- (time.north |- adaln.east) -- (adaln.east);

  \begin{pgfonlayer}{background}
    \node[draw=black, line width=0.85pt, rounded corners=13pt, minimum width=6.00cm, minimum height=6.55cm, inner sep=0pt] (outer) at (leftframe) {};
    \node[draw=black, line width=0.85pt, rounded corners=13pt, minimum width=6.00cm, minimum height=6.55cm, inner sep=0pt] (zoomframe) at (rightframe) {};
  \end{pgfonlayer}

  \draw[dashedlink] ($(block.east)+(0,0.06)$) -- ($(zoomframe.west)+(0,1.70)$);
  \draw[dashedlink] ($(block.east)+(0,-0.06)$) -- ($(zoomframe.west)+(0,-1.70)$);
\end{tikzpicture}}
  \caption{Cacheable AdaLN factorization for block diffusion. The clean prefix path is evaluated once without timestep modulation, producing a timestep-invariant cache $\mathcal{C}^{\ell}(c)$ containing attention keys/values or recurrent sequence state. At each reverse step, AdaLN$(e_t)$ modulates only the active-block continuation; the top $+$ denotes the residual add after the MLP update. This keeps prefix reuse aligned with the BDLM factorization, rather than a post-hoc approximation to a full-sequence denoiser.}
  \label{fig:adaln-factorization}
\end{figure}

This factorization is the reason the same prefix cache can be reused across reverse diffusion steps. It is independent of whether the cached state contains attention keys and values, Mamba recurrent states, or another layer-specific summary.

\end{document}